\documentclass[conference]{IEEEtran}

\usepackage{times}

\usepackage[numbers]{natbib}
\usepackage{multicol}
\usepackage[bookmarks=true]{hyperref}
\usepackage{graphicx}
\usepackage{amsmath}
\usepackage{amssymb}
\usepackage[capitalise]{cleveref}
\usepackage{xcolor}
\usepackage{algorithm}
\usepackage{algorithmic}
\usepackage{booktabs}
\usepackage{siunitx}
\newif\ifshowappendix
\showappendixtrue

\newcommand{\command}{\mathbf{u}}
\newcommand{\state}{\mathbf{x}}
\newcommand{\commandt}{\mathbf{u}(t)}
\newcommand{\knots}{\mathbf{\kappa}}
\newcommand{\statet}{\mathbf{x}(t)}
\newcommand{\commandupdate}{\Delta \mathbf{u}(t)}
\newcommand{\stateerror}{\mathbf{\tilde{x}}(t)}
\newcommand{\model}{\mathcal{M}}
\newcommand{\invmodel}{\mathcal{M}^{-1}}

\begin{document}

\title{Learning Dynamic Rope Manipulation Using Task-Level Iterative Learning Control}

\author{Krishna Suresh and Chris Atkeson\\
	Carnegie Mellon University\\
	Email: \texttt{{ksuresh2,cga}@andrew.cmu.edu}\\
\url{https://flying-knots.github.io}
}

\newcommand{\shrink}{\def\baselinestretch{0.97}\large\normalsize} 
\shrink

\maketitle

\begin{abstract}
   We introduce a Task-Level Iterative Learning Control method for dynamic manipulation of ropes. We demonstrate this method on a non-planar rope manipulation task called the flying knot. Using a single human demonstration and a simplified rope model, the method learns directly on hardware without reliance on large amounts of demonstration data or massive amounts of simulation. At each iteration, the algorithm inverts a model of the robot and rope by solving a quadratic program to propagate task-space errors into action updates. We evaluate performance across 7 different kinds of ropes, including chain, latex surgical tubing, and braided and twisted ropes, ranging in thicknesses of 7--25\,mm and densities of 0.013--0.5\,kg/m. Learning achieves a 100\% success rate within 10 trials on all ropes. Furthermore, the method can successfully transfer between most rope types in 2--5 trials. \url{https://flying-knots.github.io}
\end{abstract}

\IEEEpeerreviewmaketitle

\section{Introduction}
\begin{figure}
    \centering
    \includegraphics[width=\linewidth]{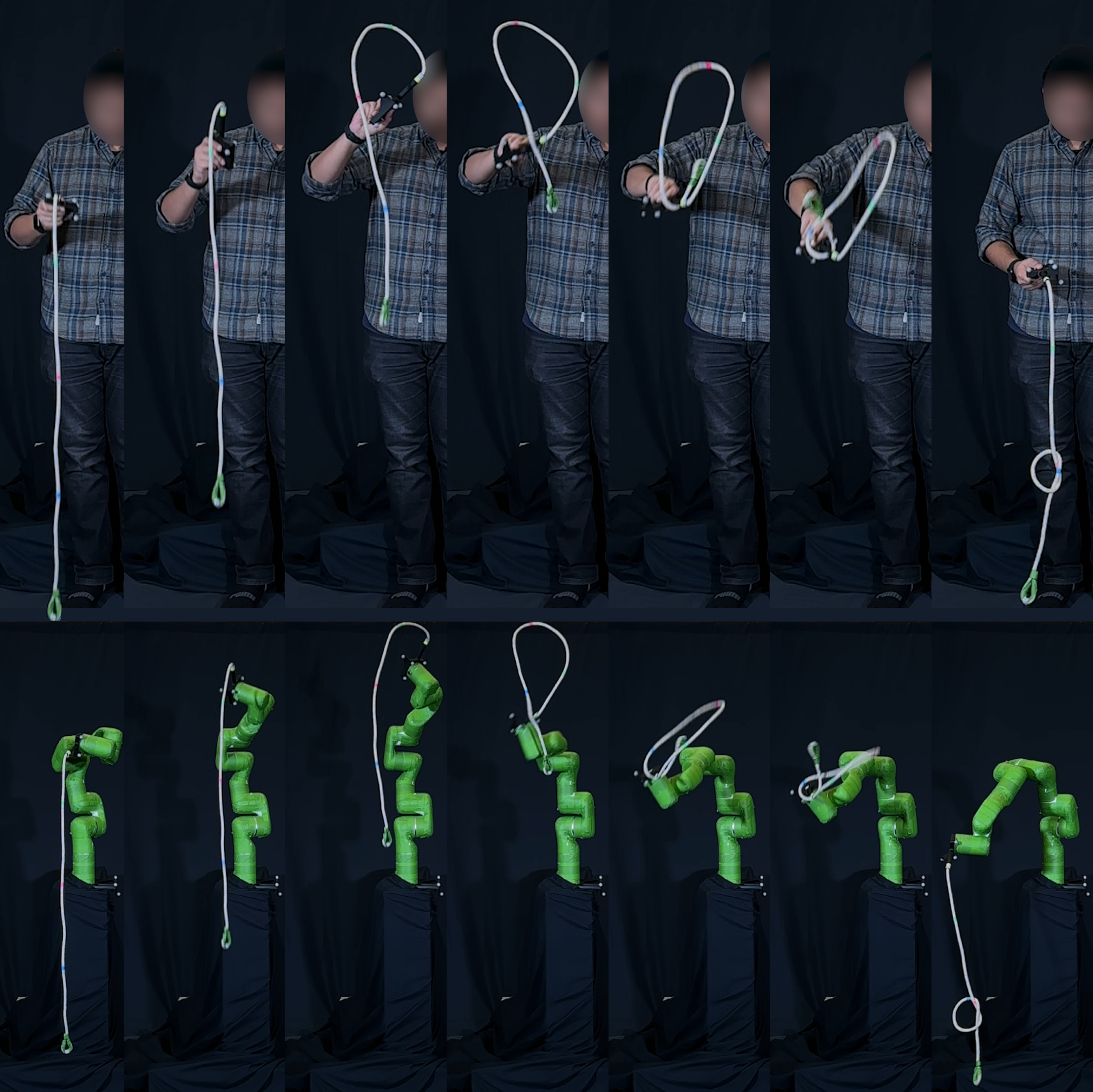}
    \vspace{-4mm}
    \caption{Stages of a flying knot by both a human and an xArm 7 robot arm over 0.56s}
    \vspace{-6mm}
    \label{fig:flying_knot}
\end{figure}

Dynamic manipulation of deformable objects is a challenging domain for both robots and humans. Deformable objects such as ropes and cloth have many unactuated degrees of freedom and are expensive to model accurately.
We enable a robot to learn the dynamic task of tying a \textit{flying knot} as seen in \cref{fig:flying_knot}. This task is performed on a rope by executing a one-handed upward and twisting motion to form a loop and then arcing to strike near the end of the rope to flip it into the loop, completing a knot.

We adapt Iterative Learning Control (ILC) \cite{1636313} to improve a command trajectory through real-world trials. ILC is a powerful tool for enabling a robot to improve its performance, as introduced by \citet{arimoto_bettering_1984}. We find that typical ILC formulations fail in deformable-object manipulation. 
Model-based iterative learning control uses an approximate system model to
convert task errors from real trials into command corrections.
When ILC is learning to track rope target motions, however, equal weighting of task errors causes learning to fail. 


Task-Level ILC uses feedback of task states in addition to or instead of robot states during learning \cite{12245}.
We leverage Task-Level ILC with two key features to enable success on dynamic rope manipulation:
\begin{itemize}
  \item \textbf{Critical point objective:} 
  Instead of weighting trajectory errors equally throughout the task, we focus the robot's attention on the error in achieving a single rope state in the trajectory. This improves learning performance and enables the transfer of the task across large parameter variations. 
   \item \textbf{Object trajectory learning:} Our approach corrects the trajectory
   of state variables of the manipulated object, which
   goes beyond the usual ILC approach of correcting the
   robot trajectory.
\end{itemize}

We demonstrate the effectiveness of Task-Level ILC for dynamic rope manipulation. We show that learning adapts to variations in rope dynamics, demonstrations, and the system model. Overall, the key contributions of this work include the following:
\begin{itemize}
\item We extend ILC to refining the trajectories of unactuated degrees of freedom of manipulated objects.
\item We introduce Task-Level ILC for
dynamic rope manipulation, demonstrating that learning typically requires a small number of real robot trials ($<10$) and transfers
across different ropes.
\item We develop a new approach to ILC, which focuses attention
on \textit{critical points} of the error history.
\end{itemize}


\section{The Flying Knot Task}
For this case study, we considered a variety of one
dimensional object manipulations, including
whipping, lassoing, throwing a rope to a target, fly casting, and attaching a rope to a distant object, such as a cleat or
tree branch by manipulating one end.
The flying knot task was chosen because it is impressive
but achievable by humans and fast robots, and has been previously explored in robotics by \citet{yamakawa_dynamic_2013}.

There are several types of flying knots. We focus on the \textit{Overhand Knot}, in which one hand manipulates the end of a rope to tie an overhand knot, without changing the grip. As seen in \cref{fig:flying_knot}, the flying knot consists of multiple stages. The rope starts hanging from the hand. The hand is moved upward and twisted to form a loop. The rope collides with itself, and the end of the rope is flipped through the loop. The knot tightens as the rope falls. Often, a weight is added to the end of the rope to make the task easier.

\subsection{Flying Knot Objective}
\label{sec:flyobj}
In this case study, we develop the idea that it is useful to define a
proxy goal for learning dynamic tasks, which we refer to as a
``\textit{critical point}''.  The use of a proxy goal can simplify the
learning task by making it easier to evaluate task success. Often the
conclusion of a dynamic task results in a crash or a state where error
information is lost. To our surprise, paying attention to errors
equally throughout task execution is often not as useful as focusing
on a particular point or phase of the task.  The selected proxy goal
can be based on an event such as the initiation of a particular
contact, on some measure of task phase, or wall clock time from some
trigger event. There are many such moments in the execution of a
flying knot: the initial state, the first instant the loop is formed,
the maximum outward swing or height of the rope end or any other part
of the rope, the rope-rope collision, the rope end passing through the
loop, or the final knotted state.

\begin{figure}
    \centering
    \includegraphics[width=\linewidth]{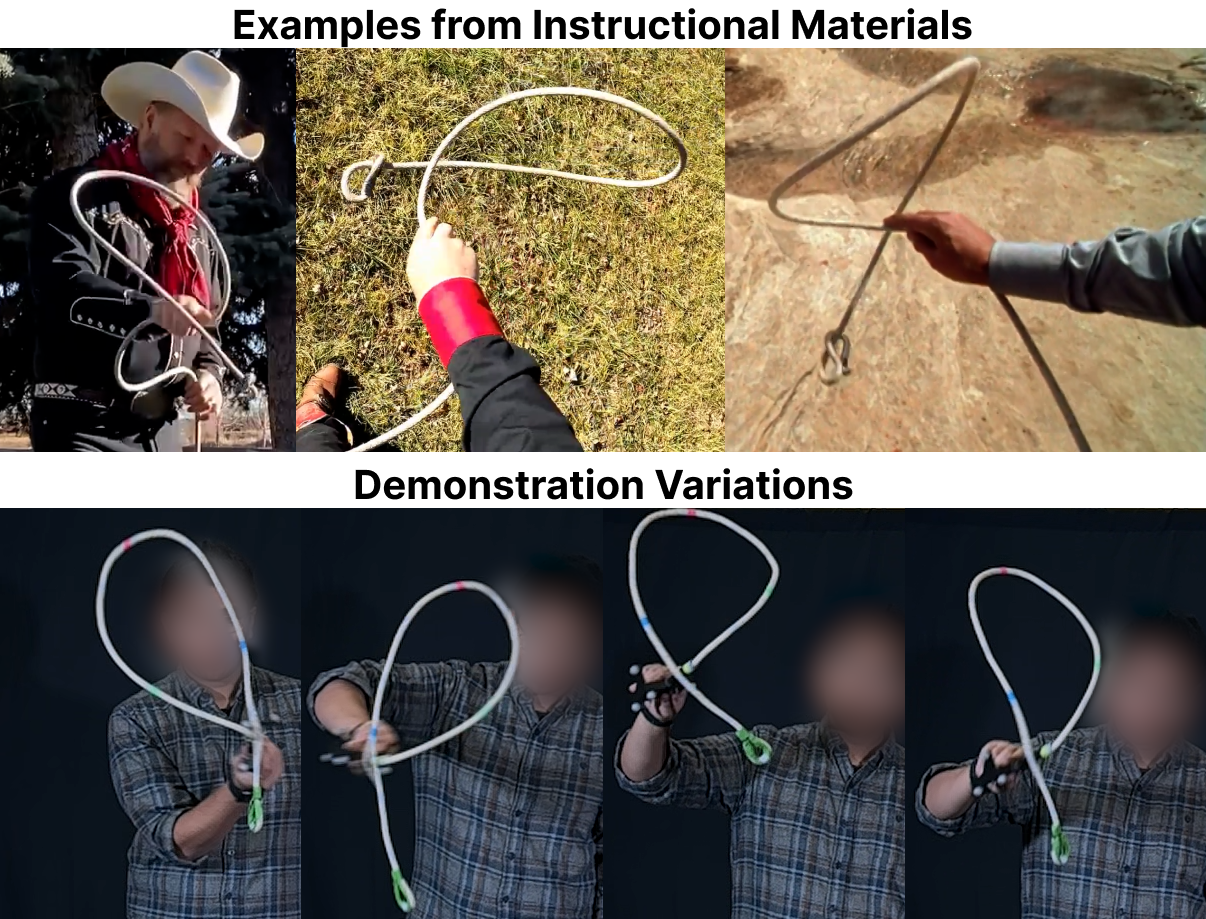}
    \vspace{-4mm}
    \caption{\textbf{Critical point} for the flying knot shown in instructional videos \cite{mink2021flyingknot, bruce2018overhandknot}. The bottom row is the \textit{critical point} state across 4 demonstration variations. The shape of the rope at collision is used as the learning objective. }
    \label{fig:demos}
    \vspace{-5mm}
\end{figure}

We chose the moment the rope collides with itself as the critical
point in this case study.  We were heavily influenced by instructional
materials which emphasized this collision state or ``strike point'' as a
crucial step, as seen in \cref{fig:demos}.  The overhand flying knot task can be
performed in several ways, but the variations share the same
qualitative topological contact event, as visualized in \cref{fig:demos},
allowing the use of a single definition of \textit{critical point}.  We also
chose this collision point because it is easy to detect and measure
and occurs before our rope tracking system gets confused by the
overlap of the rope during knot formation.  While this moment in
flying knot execution is not a perfect predictor of task success,
correctly matching this state will likely lead to a knot, so we
selected this contact initiation and the corresponding rope shape and
velocity as the \textit{critical point}.

We provide the robot with a single demonstration of the flying knot,
with the \textit{critical point} manually annotated, for an example of the rope
state at the \textit{critical point}, to provide a goal for learning the task.

\section{Related Work}
\subsection{Iterative Learning Control}
Iterative learning control is a command update algorithm in which the system iteratively executes a feedforward command in the real world and maps errors into feedforward command corrections. 
ILC leverages a system's repeatability to accumulate corrections across trials. Model-based ILC maps task errors to command updates with an \textit{inverse model} \cite{an_model-based_1988}. 

\textbf{Classical Iterative Learning Control (ILC)} studies how to improve performance on tasks that repeat over trials by updating a feedforward command based on robot trajectory errors. Early work formalized the idea of ``bettering'' robot operation via iterative error-based updates \citep{arimoto_bettering_1984}. In robotics, model-based ILC-style trajectory learning through practice was demonstrated on manipulators using nonlinear rigid-body robot models \citep{atkeson_robot_1986,an_model-based_1988}, and subsequent work analyzed convergence and design choices for robot manipulators and other repetitive systems \citep{kuc_iterative_1991,moore_iterative_1992}.

\textbf{Optimization-based ILC} or Norm-Optimal ILC formulations showed how to incorporate constraints and compute command corrections via an optimization objective. \citet{schoellig_optimization-based_2012} demonstrated aggressive quadrotor trajectory tracking by leveraging convex programs to reduce errors, and \citet{4020374} applies norm-optimal methods for real-time control of industrial gantry robots.



\textbf{Event-based ILC methods} focus learning on a subset of the task, such as the terminal states \cite{chi2015adaptive}, point-to-point \cite{8013126}, time windows \cite{1384669}, and event triggers \cite{lin2020event}. In general, placing learning attention at key moments of the task has worked in many domains, including non-robotics industrial domains such as CNC machine tools, wafer-stage motion systems, injection molding, induction motor drives, automotive braking/vehicle control, rapid thermal processing, and semi-batch chemical reactors \cite{1636313}.

\textbf{Modern learning methods} have recently been combined with ILC: using approximate simulators to propose local policy improvements \citep{abbeel_using_2006}, framing ILC through regret minimization \citep{agarwal_regret_2021}, and explaining when ILC can outperform planning with a misspecified model \citep{vemula_effectiveness_2022}. Additional work introduces ILC-style model-gradient updates in deep off-policy reinforcement learning to improve sample efficiency \citep{gurumurthy_deep_2023}. 
In our work, we leverage optimization-based and event-based techniques to focus learning on the \textit{critical points} of the task.


\subsection{Deformable Object Manipulation}
\textbf{Quasi-static deformable object manipulation}, such as the handling of rope, hair, and cloth, has been addressed in a number of works \cite{9813374, doi:10.1126/scirobotics.abd8803, zhao2026dymohairgeneralizablevolumetricdynamics} where many approaches focus on domains with strong perception and geometric priors \citep{tang_framework_2018, yoo2024ropotterroboticpotterydeformable} and imitation/representation learning for rope manipulation \cite{nair2017combining,hannus2024dynamic,huang2023act}. 

\textbf{Dynamic deformable object manipulation} focuses on faster tasks, such as whipping \cite{nah_2020}, cloth flinging \cite{chen_efficiently_2022, 5979606}, and dynamic rope knotting \citep{yamakawa_dynamic_2013}. 
Iterative Residual Policy proposes online action refinement via a learned residual dynamics model \citep{chi_iterative_2022}. Related works study learning for whip targeting, combining structured motion primitives with online adaptation or reinforcement learning \citep{xiong_online_2021,wang_learning-based_2024}.

Recent work has further explored dynamic cable and cloth behaviors with self-supervised data collection. Real2Sim2Real learning enabled planar robot casting of cables by fitting models to physical rollouts \citep{lim_real2sim2real_2022}, and subsequent systems learned to dynamically manipulate fixed-endpoint cables via arcing motions \citep{zhang_robots_2024} or free-end cables in planar settings \citep{wang_self-supervised_2024}. 

In \citet{ha2021flingbotunreasonableeffectivenessdynamic}, \citet{chi_iterative_2022}, and \citet{zhang_robots_2024}, approximate system models are learned with large-scale simulated data, but our use of model-based ILC eliminates the need for large-scale simulated data collection. 

In these works, commands are executed as feedforward trajectories and demonstrate the repeatability of dynamic deformable object manipulation. Similarly, we leverage feedforward trajectories in the flying knot task.

\cref{alg:task_level_ilc} summarizes Task-Level ILC.
\begin{algorithm}
  \caption{Task-Level Iterative Learning Control (Task-Level ILC)}
  \label{alg:task_level_ilc}
  \begin{algorithmic}[1]
    \STATE \textbf{Inputs:} initial command $\command_{0}(t)$; reference critical-point state $\state^{\text{demo}}(t_c)$; \textit{critical point} $t_c$; inverse model $\invmodel$; $K$ iterations
    \FOR{$k = 0$ \TO $K$}
      \STATE $\state_{k}(t) \leftarrow \texttt{Trial}(\command_{k}(t))$ 
      \STATE Critical-point error $\tilde{\state}_{k}(t_c) \leftarrow \mathbf{x}_{k}(t_c) - \state^{\text{demo}}(t_c)$
      \STATE Command update $\Delta \command_{k}(t) \leftarrow \invmodel\!\left(\tilde{\state}_{k}(t_c)\right)$ 
      \STATE Update $\command_{k+1}(t) \leftarrow \command_{k}(t) - \Delta \command_{k}(t)$
    \ENDFOR
  \end{algorithmic}
\end{algorithm}

\section{Method: Task-Level Iterative Learning Control}

\subsection{Iterative Learning Control}
\label{sec:method_ilc}
Iterative Learning Control (ILC) is a data-efficient learning method for repetitive tasks. ILC maps error trajectories $\stateerror$ to command trajectory updates $\commandupdate$ to iteratively refine a full sequence of commands. ILC can improve performance beyond methods that generate commands solely from a system model, since it can correct for unmodeled dynamics. 
ILC is often applied to robot trajectory-tracking problems to minimize tracking errors caused by unmodeled system dynamics (e.g., friction, cogging torques, and gear backlash and play).

ILC is well-suited when several conditions hold: the task has a fixed goal across trials, the dynamics are repeatable enough that errors persist from one trial to the next, the system can be reset to a consistent initial condition, the outcome is measureable, and the initial trajectory is close enough to the desired trajectory for the particular update operator to converge \cite{an_model-based_1988}. Model-based ILC works well when the model is good enough to get the initial trajectory close enough to the desired trajectory, and for learning to converge \cite{an_model-based_1988}.


The command trajectory $\commandt$ that ILC corrects is a function of task phase (or time) rather than a control policy, which is a function of state. 
A time-dependent feedforward command has O(N) parameters, where N is the number of samples. A feedback control policy typically has several orders of magnitude more
parameters. 
We further reduce the dimensionality of the action by representing the command trajectory with knot points of splines $\knots$.

\begin{figure}
    \centering
    \includegraphics[width=\linewidth]{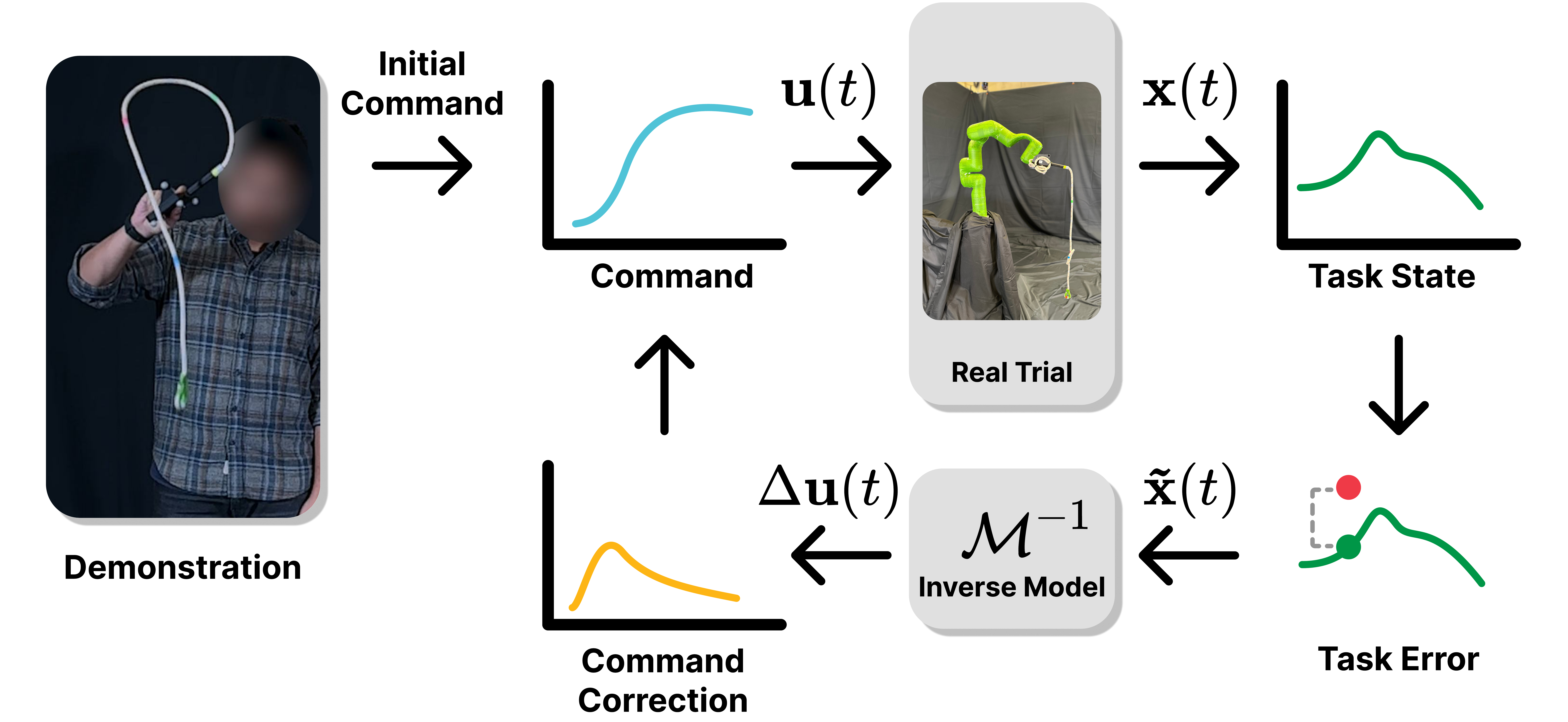}
    \vspace{-5mm}
    \caption{\textbf{Task-Level Iterative Learning Control System:} A demonstration is converted to an initial command. The command trajectory $\commandt$ is executed on the real system, and the resulting trajectory $\statet$ is measured. The task error $\stateerror$ at the \textit{critical point} is mapped through the inverse model $\invmodel$ to command trajectory corrections $\commandupdate$, which are applied to the current feedforward command, closing the learning loop.}
    \label{fig:ilc_system}
    \vspace{-6mm}
\end{figure}

Model-based ILC uses a system model $\model$ which maps commands to predictions of task performance. This system model may be inaccurate, but its gradient information accelerates learning, since knowledge of the command update direction need not be obtained from the real system. ILC inverts the system model, $\invmodel$, to map the real system's task errors into estimated command errors, which are subtracted from the current command. An overview of ILC is seen in \cref{fig:ilc_system}.
If the system model is perfect, then all errors would be removed in one step (of course, the initial command would have been perfect and there would be no errors to correct). Even when the system model is inaccurate, if the command update direction (gradient) has a component in the required command update direction, command updates accumulate and iteratively reduce task errors. 

\begin{figure*}[t]
  \centering
  \includegraphics[width=\textwidth]{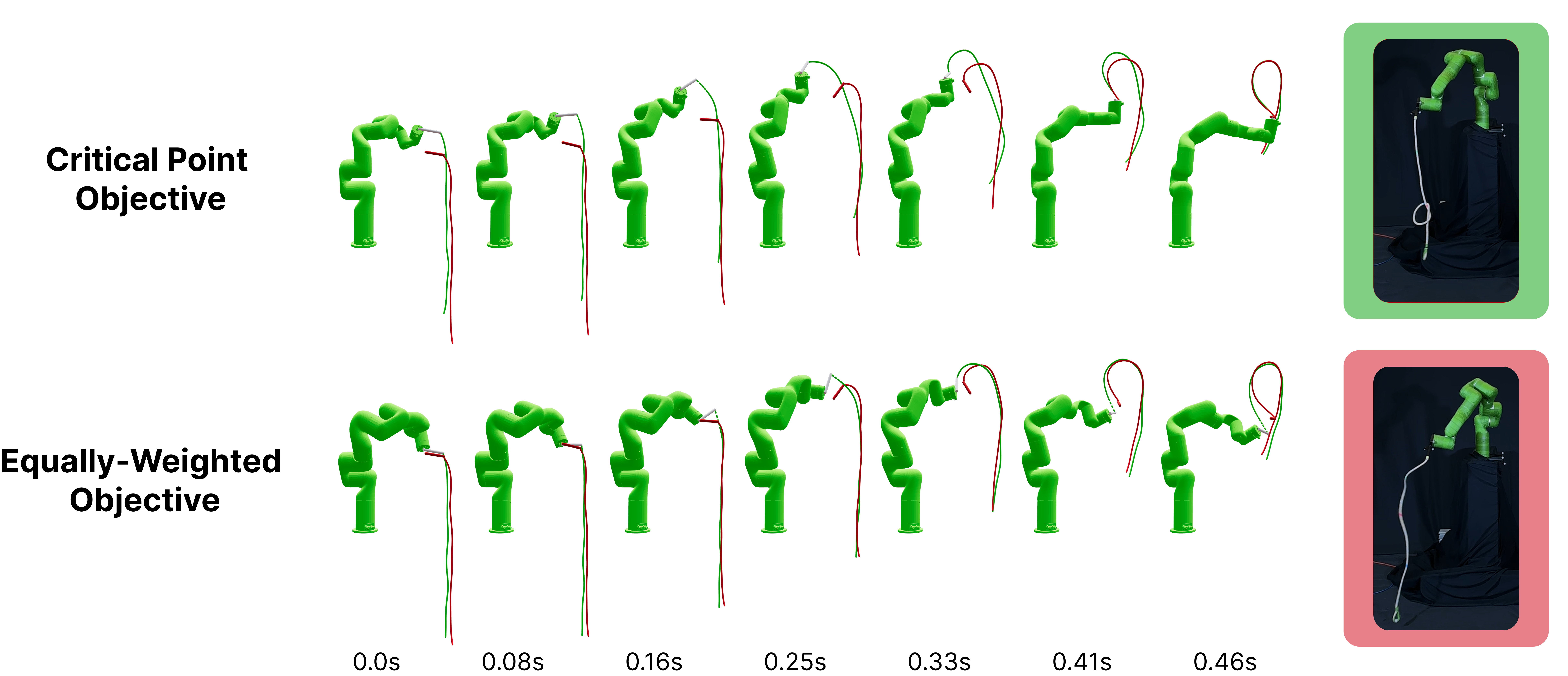}
    \vspace{-4mm}
  \caption{\textbf{Critical point vs equally-weighted objective learned commands.} Each row is a real trial after 8 iterations with the corresponding objective. The green rope is the measured state from the real trial, and the red rope is the corresponding rope state from the demonstration. The rope state at 0.46s is the \textit{critical point}. The \textit{critical point} objective results in a successful flying knot, while the equally-weighted objective results in failure.  During learning, the robot learns to deviate from the demonstration motion to reduce tracking error along the rope, which is why the demonstration rope handle position does not match the measured rope handle position.}
  \label{fig:critvseq}

    \vspace{-5mm}
\end{figure*}
\subsection{Task-Level Iterative Learning Control}
ILC is often used for robot trajectory tracking. In robot trajectory tracking, the objective and the command are related by the robot dynamics, which are often modeled accurately. Trajectory-tracking ILC usually weights errors along the trajectory equally. 

In Task-Level ILC, task errors are similarly mapped to feedforward robot command improvements. In this work, task errors are defined as deviations in the state variables of a deformable manipulated object from a goal trajectory. In addition, we explore the use of \textit{critical point} objectives, which focus learning on the error at a \textit{critical point} along the trajectory.





\subsection{Critical Point Objective}
\label{sec:critical_point}
The \textit{critical point} is defined as a key moment in the task at which all task error reduction is targeted. Let $t_c$ be the selected task phase and let $\state^{\text{demo}}(t_c)$ be the demonstrated task state. The corresponding objective minimizes the weighted error $\|\state(t_c)-\state^{\text{demo}}(t_c)\|^2_{\mathbf{Q}}$ rather than the integral of tracking error over the full trajectory.
For the flying knot task, we manually select the state of the rope at collision time $t_c$ in the demonstration as the \textit{critical point} as described in \cref{sec:flyobj}. 
Task-Level ILC attempts to generate a feedforward trajectory before the collision to match the rope state of the demonstration only at the moment of collision. Therefore, rope-tracking errors before and after the \textit{critical point} are ignored as seen in \cref{fig:critvseq}.

Eliminating task errors from the objective for times after contact also simplifies the modeling problem, as we only need to model free rope motion rather than the more complex contact physics during the collision.


\subsection{Command Parameterization}
We parameterize the feedforward robot command as 10 Bezier curves (7 joints + 3 base translation dimensions) with 8 knot points equally spaced in time: $\knots \in \mathbb{R}^{10 \times 8}$. 
The base translation dimensions ensure translation invariance of the task objective, with constraints that prevent the robot base location from moving \cref{eq:base}. 
The fixed total execution time of the command ($T$) is defined by the total length of the demonstration hand motion. 
Given $\knots$ and $T$, a Bezier curve is computed using the Bezier function $\mathcal{B}$. This spline generation of the robot's
desired trajectory command is described in more detail in Appendix~\cref{sec:bezier}.
\subsection{Robot Model}
We model the robot as a kinematic chain parametrized by robot joint angles $\mathbf{q}$. Because the robot's controller is driven by commanding
desired robot joint angles, any desired trajectory $\mathbf{q}_d(t)$ that satisfies the robot's joint position, velocity, and acceleration limits is modeled with the robot motion matching the command, $\mathbf{q_d=q}$. 
The desired trajectory is executed by the control system provided by the robot manufacturer, which mostly consists of joint-level PD servos. 
In reality, the robot motion does not exactly follow the commanded trajectory, but this error is compensated for
indirectly during task-level ILC.
Joint trajectories $\mathbf{q}(t)$ are mapped through the robot's forward kinematics $\mathcal{K}(\mathbf{q}(t))$ to determine the trajectory of the fingertip (a tube holding the rope).
While the real robot has more elements that can be modeled (e.g., link dynamics, actuation model), we find that this simple model is sufficient for ILC to improve the command. The robot model does not need to account for rope motion since the robot has joint actuators with a 100:1 gear ratio, so the rope's impact on the robot's dynamics is negligible.

\subsection{Rope Model}
We model the rope as a 3D serial chain of point masses $m$ connected by fixed-distance constraints of length $l$. Each joint in the rope has a bending stiffness $k$ and a damping coefficient $b$. Since ropes are approximately self-similar, we use a single set of $m,k,b$ parameter for all but the last rope links and joints. All the ropes have a weight attached to the end, which is approximated with a different end link mass $m_e$. Our approximate rope model has fewer degrees of freedom than the actual ropes, with only $N=11$ links. The unit scaling of these parameters is with respect to $m$, which we set to 1, resulting in only 5 model parameters ($k, b, m_e,l,N$).

\begin{figure}
    \centering
    \includegraphics[width=\linewidth]{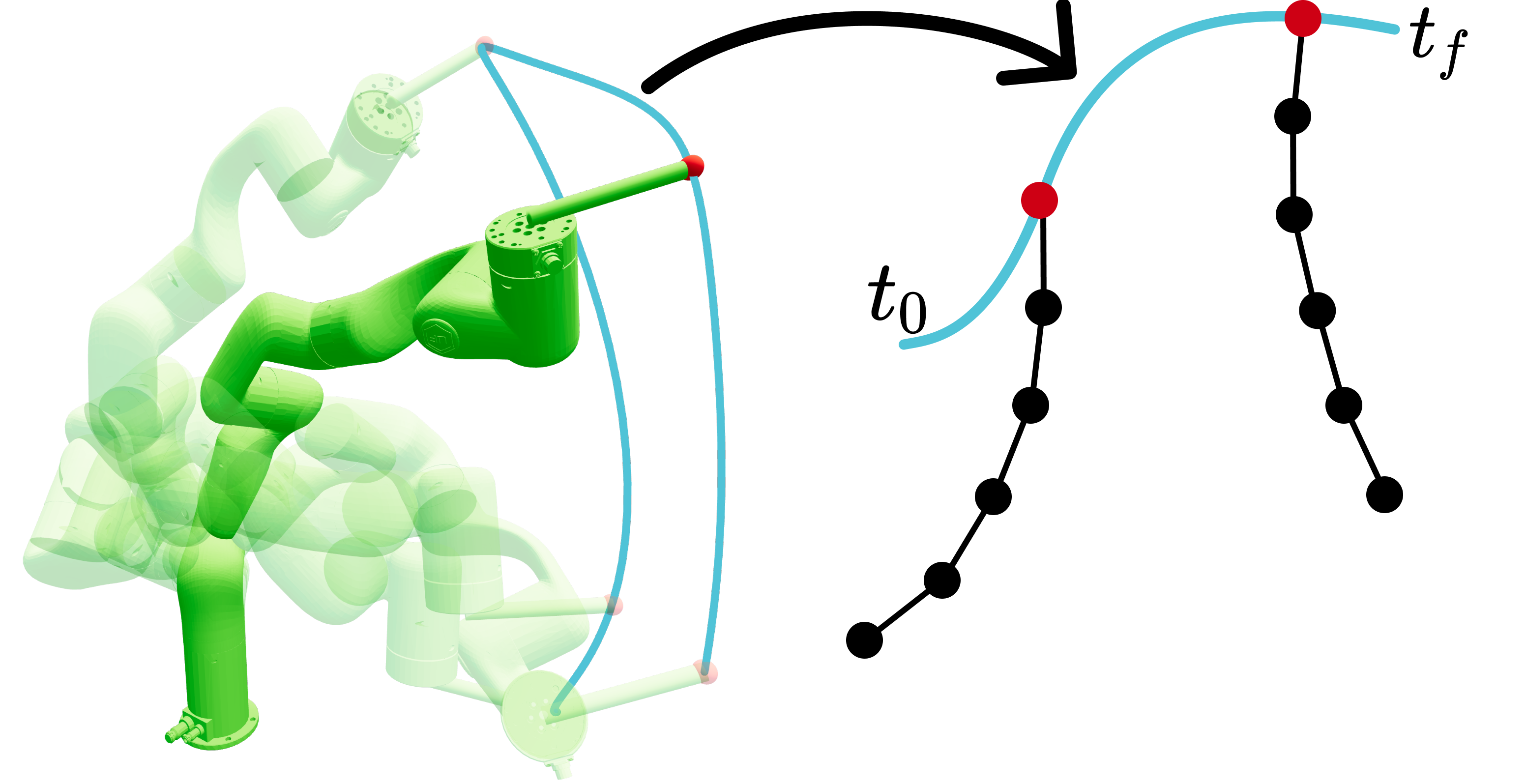}
    \caption{\textbf{Left:} Kinematic robot model at various stages of a command. The opaque robot is at the point of contact. \textbf{Right:} Graphical representation of the point mass rope model. The robot's fingertip trajectory kinematically drives the first red dot. The remaining links are bound by distance constraints, and each joint has stiffness and damping. Together, the robot and rope models define our system dynamics model. All 3D visualizations are created using Viser \cite{yi2025viser}.}
    \label{fig:system_model}
    \vspace{-4mm}
\end{figure}

The first point on the rope model is kinematically driven by the robot model's fingertip position and orientation trajectory. We can define the rope dynamics model as a function $f$, which simulates the rope state given the initial rope configuration $\mathbf{z}_0$ and a trajectory of the first link (full derivation found in Appendix~\cref{sec:dynamics_model}). This makes the overall system model, as seen in \cref{fig:system_model}, the following:
\begin{align}
   \model(\knots, \mathbf{z}_0) = f(\mathcal{K}(\mathcal{B}(\knots, T)), \mathbf{z}_0)=\hat\state(t) \label{eq:dynamics}
\end{align}
where $\hat \state$ is the model prediction of the rope state and $\knots$ are the command spline knot points. To simulate the rope dynamics, we implement a maximal coordinate variational integrator dynamics model from \citet{lavalle_linear-time_2021}. The same fixed parameter set in \cref{tab:simulation_params} is used for all rope types; we do not fit model parameters to individual ropes.

\subsection{Optimization-Based Inverse Model}
Methods for an optimization-based inverse model were introduced for trajectory tracking in \citet{schoellig_optimization-based_2012} and \citet{schollig_optimization-based_2009}. 
A Quadratic Program (QP) is formulated as a local inverse model to minimize a quadratic task objective while satisfying linear constraints by taking the measured critical-point error from the real trial and returns a command correction that is predicted to remove that error.
Previous model-based ILC methods typically use an inverse model $\invmodel$ that equally weights all parts of the trajectory, which is well-suited for trajectory tracking but is insufficient for Task-Level ILC.  
Our full inverse model QP is formulated as follows:
\begin{subequations}\label{eq:inverse_model_qp}
\begin{align}
\left(\Delta\knots_k^\star,\Delta\state_k^\star(t)\right)
&= \operatorname*{arg\,min}_{\substack{\Delta\knots\\ \Delta\statet}}
\bigl\|\Delta\state(t_c) - \tilde{\state}_k(t_c)\bigr\|_{\mathbf{Q}}^{2} \label{eq:cpobj}\\
&\qquad+ 
\sum_{t\in[t_c,T]}
  \|\Delta\knots\|_{\mathbf{Q}_{ft}, \mathbf{b}_{ft}}^{2} \label{eq:followthrough}\\
&\qquad +\sum_{t\in[0,T]}
  \|\Delta\knots\|_{\mathbf{R}}^{2} \label{eq:reg}\\
\text{s.t.}\quad
& \Delta\statet = \mathbf{M}\,\Delta\knots \label{eq:sysysy}\\
& \Delta\knots_{\text{base}} = 0 \label{eq:base}\\
& \mathbf{q}_{\text{min}} \leq \mathbf{J}_p\Delta\knots + \mathcal{B}(\knots_k) \leq \mathbf{q}_{\text{max}} \label{eq:posconst}\\
& \mathbf{\dot q}_{\text{min}} \leq \mathbf{J}_v\Delta\knots + \dot{\mathcal{B}}(\knots_k) \leq \mathbf{\dot q}_{\text{max}} \label{eq:velconst}\\
& \mathbf{\ddot q}_{\text{min}} \leq \mathbf{J}_a\Delta\knots + \ddot {\mathcal{B}}(\knots_k) \leq \mathbf{\ddot q}_{\text{max}} \label{eq:accelconst}\\
& \tau_{\text{min}} \leq \mathbf{J}_\tau\Delta\knots + \mathcal{T}(\knots_k) \leq \tau_{\text{max}}\label{eq:tauconst}
\end{align}
\end{subequations}


The QP decision variables are the command correction $\Delta\knots$ and the predicted state correction $\Delta\statet$.
Based on the measured robot motion, we simulate the rope dynamics and linearize the system model $\model$ about the current simulated rope trajectory $\hat\state_k(t)$ and current command $\knots_k$:
\begin{align}
   \mathbf{M} = \left.\frac{\partial \model}{\partial \knots}\right|_{(\hat\state_k(t),\knots_k)} \label{eq:dyn}
\end{align}

 The linearized dynamics constraint $\Delta\statet=\mathbf{M}\Delta\knots$ (\cref{eq:sysysy}) enforces a model-based match between the command and trajectory corrections.
 
 Since the ILC correction subtracts the returned correction from the current command, the first cost term penalizes the mismatch of the 
 model-predicted state correction at $t_c$ to match the measured real-system error $\tilde{\state}_k(t_c)=\state_k(t_c)-\state^{\text{demo}}(t_c)$. 
 The term \cref{eq:cpobj} is the state-error-tracking cost applied only at the \textit{critical point} $t_c$ of the demonstration. 

We apply a linearized quadratic tracking cost (\cref{eq:followthrough}) to the fingertip trajectory at $t>t_c$ to match the demonstrator's follow-through motion. The weighting matrices $\mathbf{Q}_{ft}, \mathbf{b}_{ft}$ are derived in Appendix~\cref{sec:track_hand_qp}.
 
A control cost (\cref{eq:reg}) penalizes updates to the 7 arm joints at each knot point for all $t$.

Constraint terms \crefrange{eq:posconst}{eq:tauconst} enforce position, velocity, acceleration, and torque limits on the commanded trajectory. Each has the form
\begin{equation*}
    \mathbf{x}_{\min} \;\leq\; \mathbf{J}_x\,\Delta\knots + \mathbf{B}_x(\knots_k) \;\leq\; \mathbf{x}_{\max},
\end{equation*}
where $\mathbf{B}_x(\knots_k)$ evaluates the current spline-parameterized command (or its derivative for velocity and acceleration, or its torque prediction $\mathcal{T}$ from the rigid-body inverse dynamics), and $\mathbf{J}_x$ is the linearization of that mapping with respect to the command knot points. The total quantity is bounded so that the updated command stays within the robot's joint position, velocity, acceleration, and torque limits. 

The base-translation constraint (\cref{eq:base}) prevents the 3 base-translation degrees of freedom from varying across timesteps within a single command, so the robot base stays at a fixed location throughout each trial. ILC can still shift this fixed base location between iterations.

A detailed list of all cost and constraint parameters can be found in Appendix~\cref{sec:optimization_params}. The inverse model returns the command component of the QP solution, $\invmodel(\tilde{\state}_k(t_c))=\Delta\knots_k^\star$.

After linearization, the objective is quadratic and the constraints are linear, so this inverse-model subproblem is a convex QP. We formulate the QP using the Drake optimization toolbox \cite{drake} and leverage the Clarabel Solver \cite{Clarabel_2024} to solve for the command update. 

\begin{table}[t]
    \centering
    \caption{Rope Model Parameters}
    \label{tab:simulation_params}
    \begin{tabular}{l c}
        \hline
        \textbf{Parameter} & \textbf{Value} \\
        \hline
        Stiffness & $1\times10^{5}$ \\
        Damping & $50$ \\
        Link Mass & $1$ \\
        End Mass & $5$ \\
        Simulation Timestep & $0.005$ \\
        \# Links & $11$ \\
        Link Length & $0.1$m \\
        \hline
    \end{tabular}\\
    \vspace{3mm}
    {\small The unit scaling of stiffness, damping, and end mass is defined with respect to a link mass of 1.}
    \vspace{-3mm}
\end{table}

\subsection{Demonstration}
We provide the learning system with a single demonstration of the flying knot, performed on rope 1. When evaluating learning on other rope types, the same rope 1 demonstration is used as the starting point. For a given demonstration, we track the full hand trajectory and the trajectory of the rope up to collision. We use Vicon Vantage 16 motion capture \cite{vicon_vantage_v16} to track 11 markers in 3D, which correspond to the joints in the rope model. The collision point in the demonstration is manually annotated and used as the critical point $t_c$; the algorithm does not discover this event autonomously. A search procedure is then performed to find the start and end time of the demonstration as described in Appendix~\cref{sec:demoselect}. $\mathbf{h}(t)$ is the 3D pose trajectory of the hand, $\state^{\text{demo}}(t)$ is the rope trajectory, and $t_c$ is the collision point.

\subsection{Initial Guess}
\label{sec:initial_guess}
Given a demonstration of the flying knot, we initialize learning with $\knots_0$ to track the hand fingertip trajectory $h(t)$. We solve the following trajectory optimization problem to minimize fingertip tracking error while satisfying the joint dynamics limits:
\begin{align}
\min_{\knots}\quad
& \sum_{t_i\in [0,T]}\bigl\|\mathbf{e}_h(t_i)\bigr\|_{\mathbf{W}_h}^2 + w_j\sum_{t_i\in [0,T]}\bigl\|\dddot{\mathbf{q}}(t_i)\bigr\|^2
\label{eq:ik_trajopt}\\
\text{s.t.}\quad
& \mathbf{q}(t) = \mathcal{B}(\knots, T)\nonumber\\
& \mathbf{q}_{\min}\le \mathbf{q}(t)\le \mathbf{q}_{\max} \nonumber\\
& \mathbf{\dot q}_{\min}\le \mathbf{\dot q}(t)\le \mathbf{\dot q}_{\max} \nonumber\\
& \mathbf{\ddot q}_{\min}\le \mathbf{\ddot q}(t)\le \mathbf{\ddot q}_{\max} \nonumber\\
& \dot{\mathbf{q}}(0)=\dot{\mathbf{q}}(T)=\mathbf{0} \nonumber\\
& z_{\mathrm{tip}}(\mathbf{q_0}) \ge z_{\min} \nonumber
\end{align}
where $\mathbf{q_0}$ is the initial configuration of the robot.
A detailed explanation of the hand-tracking objective $\mathbf{e}_h$, weighting $\mathbf{W}_h$, and constraints is provided in Appendix~\cref{sec:track_hand}.
The initial-guess optimization is non-convex because the hand-tracking objective depends nonlinearly on the command through the Bezier curve, robot forward kinematics, and rotation error. We therefore formulate this trajectory optimization problem using the Drake toolbox \cite{drake} and solve using the SNOPT nonlinear solver \cite{snopt}.

\begin{figure}
    \centering
    \includegraphics[width=\linewidth]{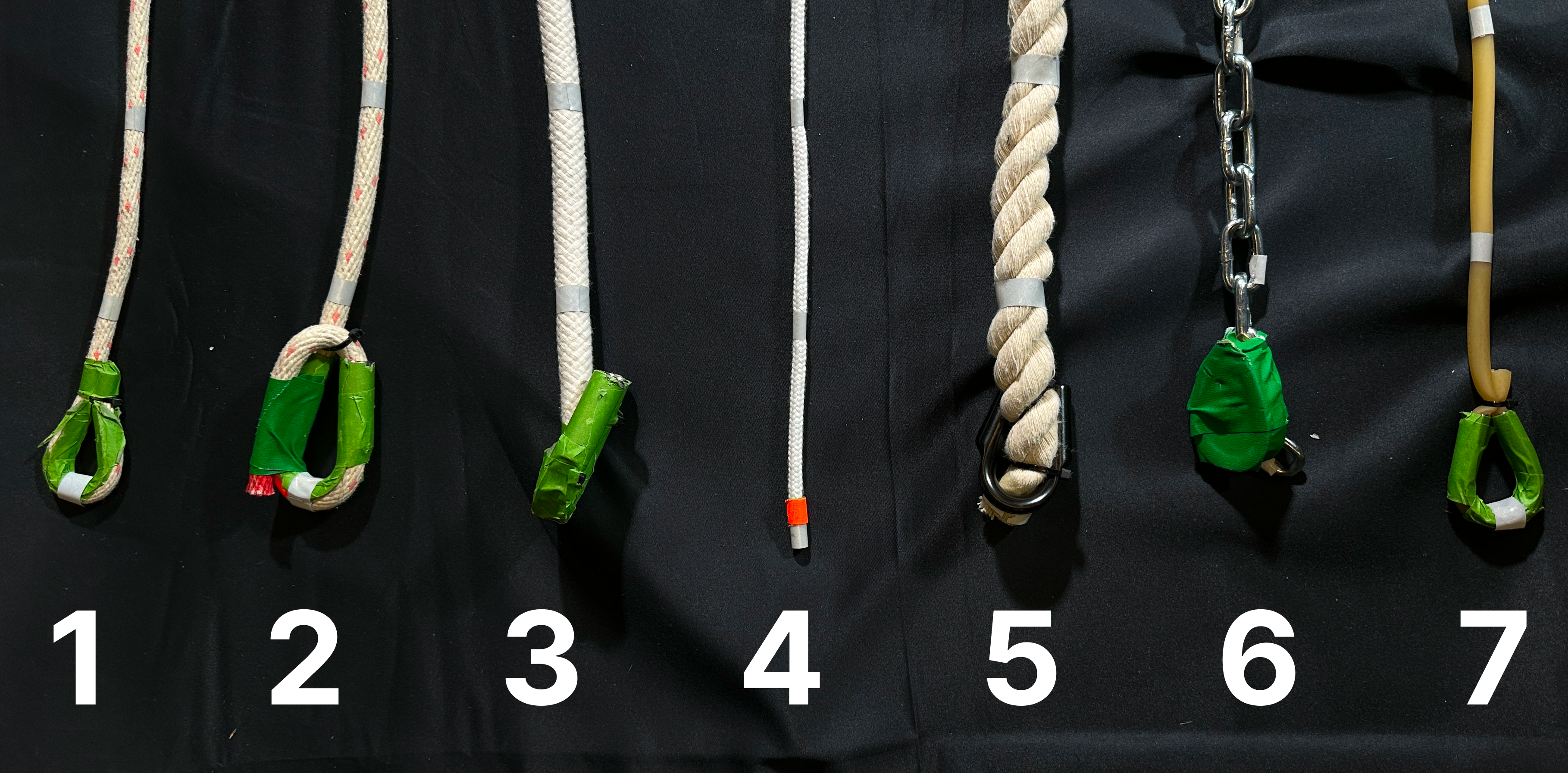}
    \vspace{-3mm}
    \caption{7 different rope types used for evaluation, including 4 braided ropes of various thickness and mechanical properties, a large twisted rope, a chain, and latex surgical tubing (very elastic); ranging in thicknesses of 7--25\,mm and densities of 0.013--0.5\,kg/m. Each rope has a mass affixed to aid in the formation of the knot.}
    \label{fig:ropes}
    \vspace{-4mm}
\end{figure}

\section{Evaluation}

We evaluate the learning algorithm's performance across 7 rope types (\cref{fig:ropes}) and provide 4 variations of the flying knot demonstration (\cref{fig:demos}). We define a successful flying knot as achieving a topological overhand knot in a rope hanging down at the end of motion (e.g. a knot being tied then untied during motion is a failure). Given a flying-knot demonstration, we compare the success of our Task-Level ILC approach against two common approaches: direct tracking of the human demonstrator (hand motion) and equally-weighted ILC learning of task progression (rope motion). The resulting commands are tested for robustness with 40 trials. We then evaluate the transfer of a learned command across rope types and, finally, assess the system's learning robustness with respect to model parameters.





\subsection{Experiment Setup}
We conduct flying knot experiments using the xArm 7 robot arm \cite{ufactory_xarm7}. Both command execution and robot motion measurement occur at 250Hz. 
Commands that exceed the robot's joint dynamics limits in position, velocity, acceleration and torque result in mid-motion faults and task failure. 

Each rope is marked with 11 rings of 12\,mm wide retroreflective tape wrapped around the rope evenly spaced 10\,cm apart along its length. These markers are tracked with a Vicon Vantage 16 motion capture system \cite{vicon_vantage_v16} at 200Hz. Tracking of rope markers often fails once the rope-rope collision occurs due to occlusion and marker misidentification across the different cameras, resulting in bad 3D reconstructions. Hand position is also tracked using 4 motion-capture markers. All ropes are 1.1 meters long.
The rope is always started in a static state hanging down from the fingertip, though some types of ropes are stiff enough to have curved resting states. 
\begin{figure}
    \centering
    \includegraphics[width=0.85\linewidth]{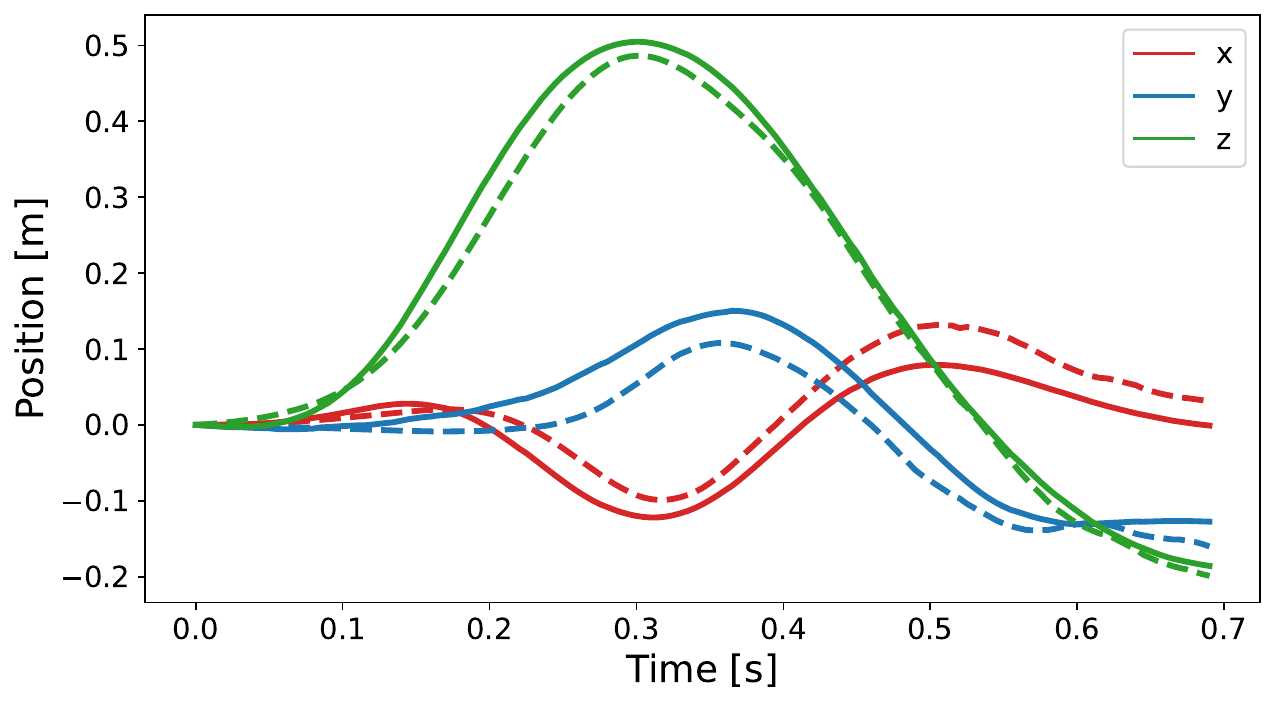}
    \caption{\textbf{Initial command fingertip trajectory}: positions of the demonstration fingertip trajectory are dashed, and the robot's commanded initial attempt to track the demonstration is solid. The robot fails to track the demonstration motion due to kinematic and dynamic constraints.}
    \label{fig:dmeo_track}
\end{figure}

\subsection{Demonstration Hand Tracking}
For each demonstration type, we apply our initial-guess generation procedure, as described in \cref{sec:initial_guess}, to assess how well the robot performs the task solely by following the demonstrated hand motion. In every rope-demonstration type for rope 1, the robot is unable to tie the flying knot when attempting to follow the demonstration trajectory. However, for certain ropes, such as 4 and 7, following the demonstration nearly achieves a flying knot, even though the demonstration provided was using rope 1.

Generating a command to exactly track the demonstrated hand motion is not possible due to the limits of the robot's joint ranges, maximum velocities, accelerations, and torques, as well as the differences between human and robot morphology. An example demonstrated fingertip trajectory and the corresponding initial attempt by the robot are visualized in \cref{fig:dmeo_track}. While the robot can capture the demonstrator's overall motion, it cannot match it exactly, resulting in failures when attempting to tie the flying knot.



\subsection{Weighted Error Objectives}
\label{sec:critical_point_obj}

We compare the performance of the learning algorithm under two objectives: the \textit{critical point} objective (ours) and the equally-weighted objective. The \textit{critical point} objective, which focuses on errors at one point in the trajectory, is described in \cref{sec:critical_point}. The equally-weighted objective weights rope tracking errors equally along the pre-collision trajectory and is commonly found in trajectory tracking and behavior cloning. 

As seen in \cref{fig:critvseq}, the critical-point objective aligns the rope state at a single point, at the collision, allowing for different rope motion beforehand, and results in success.

\begin{figure}
    \centering
    \includegraphics[width=\linewidth]{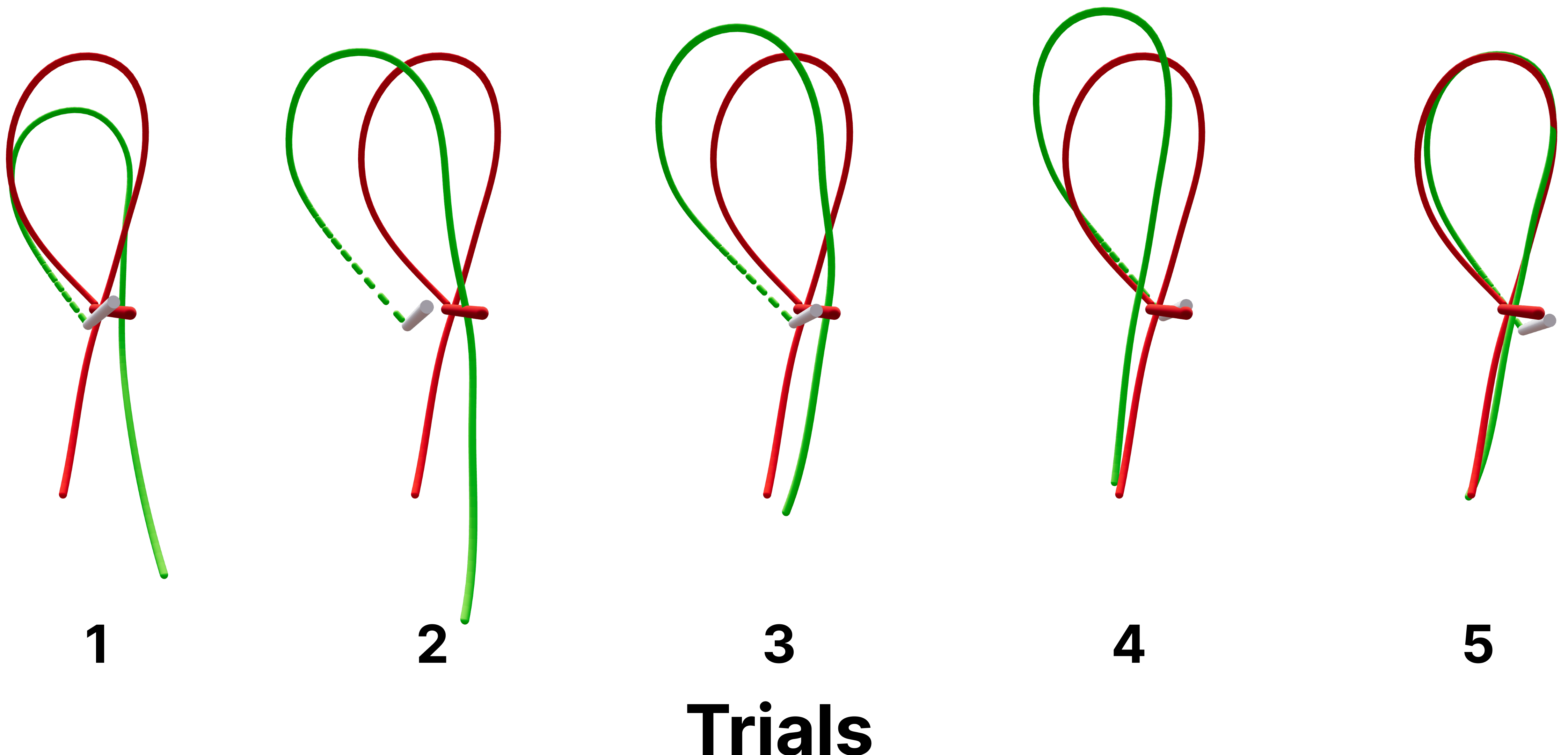}
    \vspace{-3mm}
    \caption{\textbf{Rope configuration at the \textit{critical point} over learning iterations:} Real rope (green) and demonstration rope (red) states for 5 iterations of Task-Level ILC where trial 5 resulted in a flying knot.}
    \vspace{-1mm}
    \label{fig:iters}
\end{figure}

Overall, we find that the \textit{critical point} objective is crucial to task success. The equal weighting objective increases errors at the critical point, compared to the critical point objective, to reduce errors at earlier parts of the trajectory that matter less for task success. Over trials, the error at the critical point is reduced as shown in \cref{fig:iters}. This error difference at the \textit{critical point} for equally-weighted learning causes failure.

\begin{figure*}[t]
  \centering
  \includegraphics[width=\textwidth]{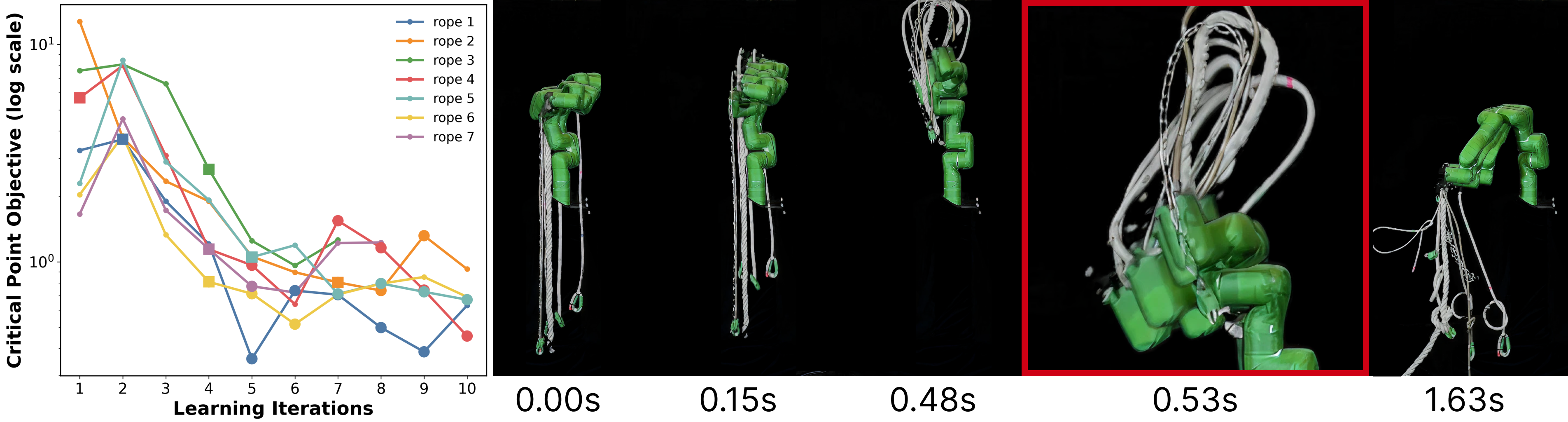}
  \caption{\textbf{Left:} \textit{Critical point} objective for 7 rope types over 10 iterations (rope 3 and 7 end early due to marker tracking failures). Squares represent the first successful flying knot, and every large solid dot represents a subsequent successful knot. \textbf{Right: } Execution on the real robot of successful flying knot execution on 7 rope types overlaid. The red highlighted frame is the \textit{critical point}.}
  \label{fig:all_rope_sequence}
\end{figure*}

\subsection{Learned Command Success Rate}
\label{sec:learned_command_success}
Once a successful trajectory is learned, the robot has a 100\% success rate with the learned command. We evaluated this success rate by repeating 40 trials of a learned command for each condition. Execution on the real robot is performed immediately after learning, and longer-term changes to the system dynamics, such as rope wear or breaking in (changes to stiffness and damping), and robot motor temperature, might require additional learning iterations.
Continued learning can degrade the command, so learning is stopped after the first success (see \cref{sec:command_learning} for discussion of high-frequency error amplification over iterations).

\subsection{Learning Across Rope Types}
We evaluate the learning of the flying knot across 7 different ropes (as shown in \cref{fig:ropes}) to demonstrate the robustness of Task-Level ILC to varying system dynamics. A detailed list of parameters of each rope can be found in Appendix~\cref{sec:rope_params}. In general, ropes span a thickness of 7--25\,mm, a density of 0.013--0.5\,kg/m, and are composed of materials such as cotton, polyester, steel, and latex.
All ropes have a mass fixed to the end with masses selected to allow the human demonstrator to execute the flying knot.



Task-Level ILC successfully learns the flying knot on all 7 rope types within 10 trials, starting from the same single rope 1 demonstration in all cases. The successful flying knot commands are shown in \cref{fig:all_rope_sequence}. Each command starts the rope in a different initial state and performs the upward twisting motion at different speeds. Once the critical moment is reached, all ropes form the same loop shape and allow the knot to be formed. The differences in learned commands show that variation in the rope dynamics requires variation in the command. 
As seen in \cref{fig:all_rope_sequence}, the learning algorithm succeeds in fewer trials for some rope types (3, 6, 7) and requires more iterations for others (2, 5), but does not exceed 10 trials on any rope. Overall, the learning process reduces the cost of the \textit{critical point} objective across trials; however, because we are not performing a line search on the real-system objective, the cost can increase after a poor command update, as with rope 4. Furthermore, repeated updates after success can lead to failure, as in trial 9 of rope 2; model errors are well known to cause ILC to slowly amplify high-frequency errors over iterations \citep{atkeson_robot_1986}. 

%



\subsection{Learning Across Demonstration Variation}
\begin{figure}
    \centering
    \includegraphics[width=0.75\linewidth]{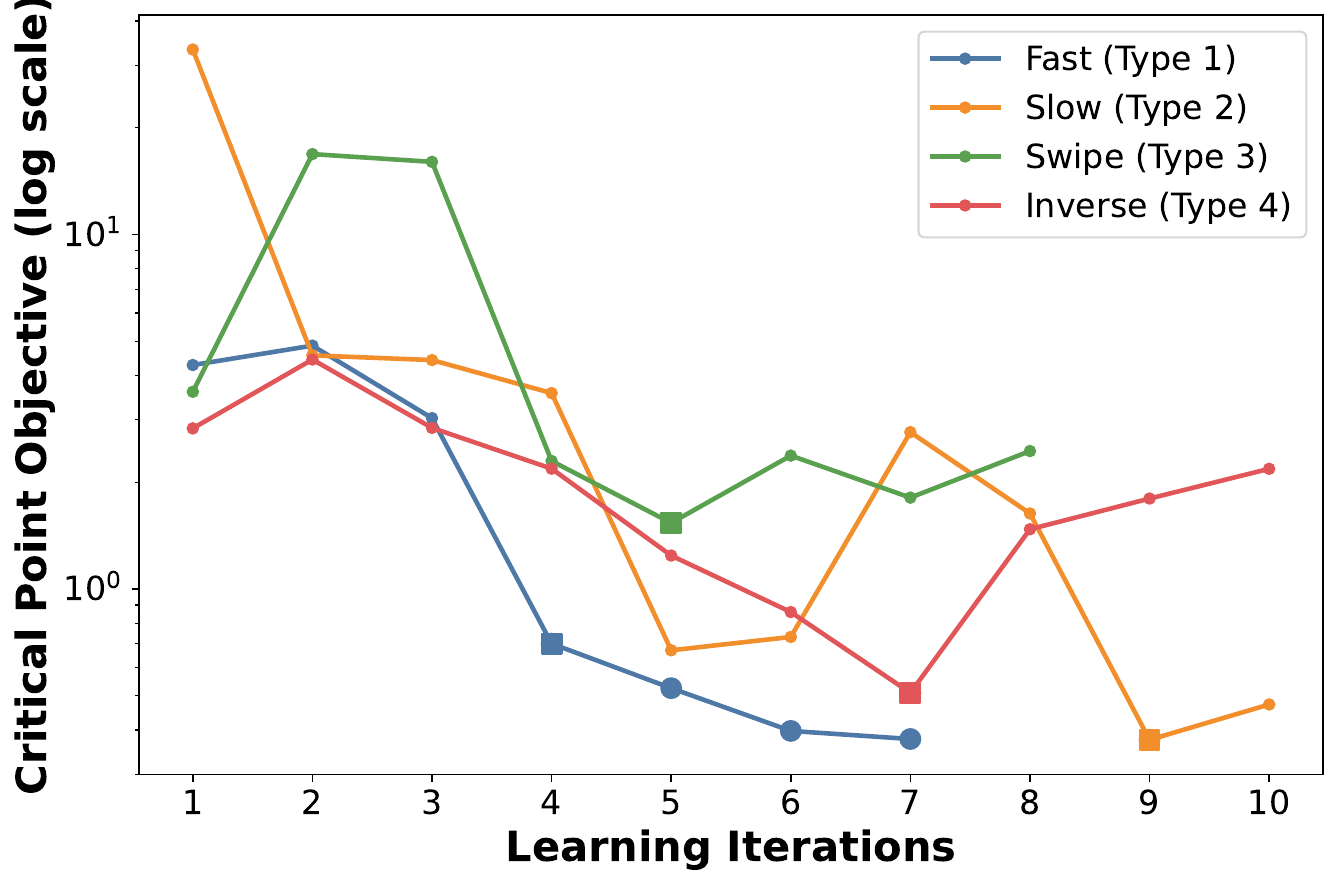}
    \caption{Learning cost over trials for 4 demonstration types. Large solid dots represent times when the flying knot succeeded, and squares are the first successful command. Learning on Demo Types 1 and 3 is truncated due to rope tracking failures.}
    \label{fig:4success}
    \vspace{-4mm}
\end{figure}

For rope 1, we test whether variations in the demonstration impact learning with the 4 different flying knot demonstration types seen in \cref{fig:demos} (Fast, Slow, Swipe, and Inverse). 
The total demonstration time lengths vary, ranging from 0.69s to 1.04s. The learning algorithm learns the flying knot in all 4 cases.  \cref{fig:4success} shows the learning progression, with Demo Type 2 requiring the most trials. 

\subsection{Transfer of Learned Command}

\begin{figure}
    \centering
    \vspace{-7mm}
    \includegraphics[width=0.75\linewidth]{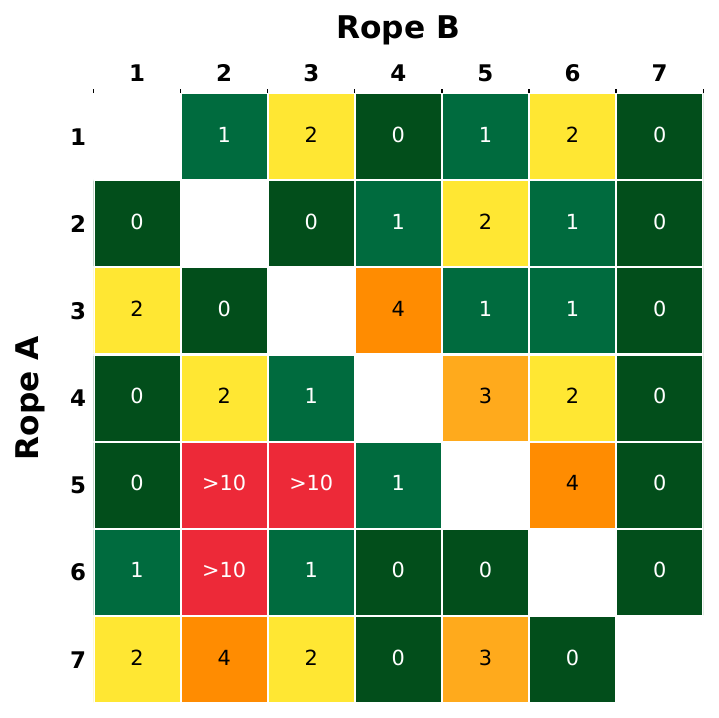}
    \vspace{-2mm}
    \caption{Number of trials to transfer a successful command from rope A to rope B. Red cells are for transfer that did not succeed within 10 trials.}
    \label{fig:transfer_learning}
    \vspace{-2mm}
\end{figure}

Given a demonstration on rope 1 and a successful command on all 7 ropes, we attempt to transfer the learned commands between rope types. The successful command on rope A initiates learning on rope B and has a maximum of 10 trials to improve. A grid of the number of trials required to adapt the command to rope B is shown in \cref{fig:transfer_learning}. For the majority of transfers, the number of trials is larger than zero, showing that the dynamics of the ropes are different enough to require different commands. For rope 7 (9mm latex surgical tubing), no additional learning is required, indicating that rope 7 is more robust to command variations. Additionally, learning fails when commands from ropes 5 and 6 are transferred to ropes 2 and 3. In all three cases, learning iteratively reduces the objective but requires larger adjustments and cannot fully correct the command within the 10 allotted trials.

\begin{table}[b]
    \vspace{-3mm}
    \centering
    \caption{Effect of model parameters on learning performance.}
    \label{tab:stiffness_endmass}
    \begin{tabular}{cc|c}
        \toprule
        Stiffness & End Mass & Trials Until Success \\
        \midrule
        $\mathbf{1\times10^{5}}$ & \textbf{5} & \textbf{4} \\
        $1\times10^{4}$ & 5      & 4   \\
        $1\times10^{3}$ & 5      & 4   \\
        $1\times10^{2}$ & 5      & $>10$ \\
        $1\times10^{5}$ & 0.05   & 8   \\
        $1\times10^{5}$ & 0.005  & 6   \\
        \bottomrule
    \end{tabular}
    \vspace{-3mm}
\end{table}

\subsection{Sensitivity to Model Parameters}
\label{sec:model_sensitivity}
ILC can make command corrections in a sample-efficient manner by leveraging a system model. We evaluate the effects of changes to the model parameters by learning on rope 1 with a single demonstration, while varying the system model stiffness $k$ and end mass $m_e$. These parameters were selected for their dominant impact on the model dynamics. In \cref{tab:stiffness_endmass}, we evaluate learning performance with different rope model parameters 
(bold represents the default parameters used across all experiments). 
For order-of-magnitude changes in the model parameters, learning can still succeed. There are cases in which learning fails, such as when stiffness is too low, in which a single command update places the trials in an unrecoverable execution regime. Similarly, for runs with low end-mass values, updates after a successful knot quickly drive the command away from success.




\section{Discussion}
\subsection{Learning on the Real System}
Task-Level ILC enables task learning on real systems in a sample-efficient manner, using approximate system models to convert real-system errors into command updates. In simulation-based learning methods, such as policy learning on a simulated system model and Sim2Real transfer with domain randomization, issues often arise due to gaps between the system model and the real system. Model learning attempts to overcome this by adapting the model used for policy optimization with real data, but often suffers from limitations of the model structure. Domain randomization builds robustness into the policy by optimizing over a range of models. Policy optimization with domain randomization is a worst-case robust control design procedure, where performance is degraded in all situations relative to the performance with an accurate model for that situation.

Learning directly on the real system eliminates these issues by allowing the real system to be captured in the learning process. The models used for learning do not need to accurately predict future system states, as illustrated by the large discrepancies between the model predictions and the real rope, yet learning still occurs. These approximate models still provide sufficiently accurate gradient information. As shown in \cref{sec:model_sensitivity}, the learning process is robust to large variations in the parameters, eliminating challenges in accurate model parameter estimation and domain-randomization.


\subsection{Critical Points for Task-Level Learning}
In evaluating the learning system on the flying knot task, we demonstrate that a \textit{critical point} enables learning, robustness, and transfer. In \cref{sec:critical_point_obj}, we show that an equally-weighted objective failed to achieve the task because errors at the \textit{critical point} are larger than with a \textit{critical point objective} (as visualized in \cref{fig:critvseq}). 

\textit{Would critical point objectives improve Behavior Cloning (BC)?}
BC achieves tasks not by iteratively improving performance but instead by capturing the task distribution with a range of demonstrations. Direct teleoperation of the robot is often required for BC to ensure real system dynamics are captured in demonstrations. In BC, task transfer often struggles across domains and robots. \citet{hu_rac_2025} addresses some of these challenges by leveraging an instructor to guide learning toward subgoals. One can view the subgoals as
critical points.


\subsection{Command Learning}
\label{sec:command_learning}
For tying a flying knot, we are learning a feedforward command. Repeated execution of commands results in reproducible behavior. By learning a feedforward command, the learning problem is reduced to searching for commands as a function of time ($\command(t)$) rather than a general feedback policy ($\command(\state)$). We further reduce the dimensionality by representing $\commandt$ with a set of knot points $\knots$. This reduction in the dimensionality of the learning problem comes at a cost of the approach being unable to 
handle unstable systems or situations with large random perturbations.
ILC can be applied to unstable systems that have been stabilized \cite{schoellig_optimization-based_2012}. 

ILC can suffer from high frequency instabilities. Mechanical systems with energy loss are inherently low-pass filters. Inverse models of these systems become high-pass filters. Learning using an inverse model as the learning operator across trials amplifies high-frequency modeling inaccuracies \citep{atkeson_robot_1986}. 

We parameterize the command trajectory using knot points of splines ($\knots$) producing motions that are smooth and continuous. This reduces the effect of high frequency instabilities.


Manipulation of deformable objects is well-suited for command learning, as errors are often repeatable, and with energy loss, the system dynamics are usually stable.


\subsection{Limitations and Future Work}
Our learning system is robust to changes in rope dynamics, different demonstrations, and model parameters. Further robustness evaluations can assess changes in rope length, non-uniform density, and end mass. 

Our current method has several main failure modes that degrade learning. First, because the system model does not account for collisions, rope collisions with the robot body or the rope handle can cause the inverse model to apply unrecoverable command corrections. Second, the selected \textit{critical point} of the rope collision is sometimes insufficient for fully completing the flying knot, as the rope may tie the knot and execute a follow-through motion, which unties the knot before the rope settles. Similarly, lighter ropes can get caught on the finger surface and be unable to tighten the knot. Third, marker tracking is only accurate up to rope collision, and since we set the \textit{critical point} to be at a fixed time point from the start of motion, if the rope contact happens early, rope state estimation may fail. Lastly, learning can diverge and get stuck in a local minimum. Learning divergence occurs more frequently when task errors are too large at the \textit{critical point}, and the linearized model is not accurate enough.

Avenues for future research include autonomous selection of \textit{critical points} from demonstrations or instructional materials, identification of model structures for a given task, and learning from lower-fidelity demonstrations/measurements. Specifying the \textit{critical point} as the rope collision point is a crucial input to our learning approach, and the autonomous identification of these \textit{critical points} will enable a robot to learn efficiently. Potential selection criteria for \textit{critical points} could include: human instruction, changes in contact state; dynamical extrema (position, velocity, acceleration, jerk, curvature, etc.); closest/furthest approach points; dynamic bifurcations/branch points; and dynamical funneling points.

\section{Conclusion} 
\label{sec:conclusion}
We demonstrate the success of Task-Level Iterative Learning Control for dynamic rope manipulation by learning to tie overhand flying knots in less than 10 trials. We introduce the notion of a critical-point objective and demonstrate task-level learning of unactuated system dynamics. Given a single demonstration of the task, our learning system learns to tie a flying knot across 7 different rope types and 4 demonstration variations. We show that a learned command achieves a 100\% success rate and transfers across most rope types in 2--5 trials. 


\bibliographystyle{plainnat}
\bibliography{references}

\ifshowappendix
\clearpage
\appendix

\subsection{Inverse Model Additional Details}
\subsubsection{Inverse Model Parameters}
\label{sec:optimization_params}
\begin{table}[t]
\centering
\caption{Inverse Model Parameters}
\label{tab:invmodelparams}
\begin{tabular}{l c}
\toprule
\textbf{Parameter} & \textbf{Value} \\
\midrule
$w_{\text{control}}$ & $\;0.5\;$ \\
$w_{\text{critical pos}}$ & $\;25\;$ \\
$w_{\text{critical vel}}$ & $\;0.00375\;$ \\
$w_{pc}$ & $\;100\;$ \\
$w_{vc}$ & $\;0.1\;$ \\
$w_{Rc}$ & $\;5\;$ \\
$w_{pft}$ & $\;1\;$ \\
$w_{vft}$ & $\;0.1\;$ \\
$w_{Rft}$ & $\;0.1\;$ \\
$w_{\text{ft velocity}}$ & $\;0.5\;$ \\
$\mathbf{q_{\min}}$ & $[\;-6.28, -1.8, -6.28, -0.19, -6.28, -1.69, -6.28\;]$ \\
$\mathbf{q_{\max}}$ & $[\;6.28, 1.9, 6.28, 3.92, 6.28, 3.14, 6.28\;]$ \\
$\mathbf{\dot q_{\min}}$ & $-[3.14, 3.14, 3.14, 3.14, 3.14, 3.14, 3.14]$ \\
$\mathbf{\dot q_{\max}}$ & $[3.14, 3.14, 3.14, 3.14, 3.14, 3.14, 3.14]$ \\
$\mathbf{\ddot q_{\min}}$ & $-[100, 100, 100, 100, 100, 100, 100]$ \\
$\mathbf{\ddot q_{\max}}$ & $[100, 100, 100, 100, 100, 100, 100]$ \\
$\mathbf{\tau_{\min}}$ & $-[130, 130, 40, 40, 40, 20, 20]$ \\
$\mathbf{\tau_{\max}}$ & $[130, 130, 40, 40, 40, 20, 20]$ \\
\bottomrule
\end{tabular}
\end{table}

We use the shorthand $\|\mathbf{a}\|_{\mathbf{W}}^2 := \mathbf{a}^\top \mathbf{W}\mathbf{a}$. The critical-point objective weights the rope-marker position and velocity errors at $t_c$ with a diagonal matrix
\begin{align*}
    \mathbf{Q} := \operatorname{diag}\!\left(w_{\text{critical pos}}\mathbf{I}_{3N},\; w_{\text{critical vel}}\mathbf{I}_{3N}\right),
\end{align*}
where $N$ is the number of rope markers (links). In general, $w$ is a diagonal cost element for a cost matrix. The control-update regularizer is $\|\Delta\knots\|_{\mathbf{R}}^2$ with a diagonal $\mathbf{R}$ applying $w_{\text{control}}$ to the 7 arm-joint update variables. For the follow-through, we penalize the end-effector tracking error in \cref{sec:track_hand} with a time-varying diagonal weight matrix
\begin{align*}
    \mathbf{W}(t) := \operatorname{diag}\!\bigl(w_p(t)\mathbf{I}_3,\; w_R(t)\mathbf{I}_3,\; w_v(t)\mathbf{I}_3\bigr),
\end{align*}
using $(w_p,w_R,w_v)=(w_{pc},w_{Rc},w_{vc})$ at $t=t_c$ and $(w_p,w_R,w_v)=(w_{pft},w_{Rft},w_{vft})$ for $t>t_c$. The parameter $\mathbf{q}$ corresponds to the seven robot joint angles, with minimum and maximum constraints imposed on angles, velocities, and accelerations. $\tau$ is the predicted joint torque as given by the inverse dynamics of a full robot model. 
All values used in our experiments are listed in \cref{tab:invmodelparams}.

\subsubsection{QP Hand Tracking Objective}
\label{sec:track_hand_qp}
For $t \in [t_c, T]$, we encourage the robot to match the demonstrator's follow-through motion by penalizing the end-effector tracking error. The follow-through reference is continuously matched to the robot motion at $t_c$, avoiding discontinuities that would alter the tracking cost. Let $\mathbf{e}_{ft}(t;\knots)$ denote the end-effector error vector defined in \cref{sec:track_hand}; this error depends on $\knots$ nonlinearly through the Bezier spline $\mathcal{B}(\knots, T)$ and the robot forward kinematics. We linearize $\mathbf{e}_{ft}$ about the current knot points $\knots_k$ with respect to a small command update $\Delta\knots$:
\begin{align}
    \mathbf{e}_{ft}\bigl(t;\knots_k + \Delta\knots\bigr)
    &\approx
    \mathbf{e}_{ft}\bigl(t;\knots_k\bigr) + \mathbf{J}_{ft}(t)\,\Delta\knots, \\
    \mathbf{J}_{ft}(t) &:= \left.\frac{\partial \mathbf{e}_{ft}(t;\knots)}{\partial \knots}\right|_{\knots=\knots_k}.
\end{align}
For notational simplicity, we write $\mathbf{e}_{ft}(t) := \mathbf{e}_{ft}(t;\knots_k)$. We apply a weighted least-squares cost
\begin{align}
    &\bigl\|\mathbf{e}_{ft}\bigl(t;\knots_k + \Delta\knots\bigr)\bigr\|_{\mathbf{W}(t)}^{2} \nonumber\\
    &\qquad\approx
    \bigl\|\mathbf{e}_{ft}(t) + \mathbf{J}_{ft}(t)\,\Delta\knots\bigr\|_{\mathbf{W}(t)}^{2}.
\end{align}
Expanding and dropping the constant term $\|\mathbf{e}_{ft}(t)\|_{\mathbf{W}(t)}^{2}$ yields a quadratic function of the command update:
\begin{align}
    \|\Delta\knots\|_{\mathbf{Q}_{ft}(t),\mathbf{b}_{ft}(t)}^{2}
    &:= \Delta\knots^\top \mathbf{Q}_{ft}(t)\,\Delta\knots \nonumber\\
    &\quad + \mathbf{b}_{ft}(t)^\top\Delta\knots, \\
    \mathbf{Q}_{ft}(t) &= \mathbf{J}_{ft}(t)^\top \mathbf{W}(t)\mathbf{J}_{ft}(t), \\
    \mathbf{b}_{ft}(t) &= 2\,\mathbf{J}_{ft}(t)^\top \mathbf{W}(t)\mathbf{e}_{ft}(t).
\end{align}

\begin{table*}[t]
    \centering
    \caption{List of Rope parameters}
    \label{tab:rope_id}
    \small
    \setlength{\tabcolsep}{6pt}
    \begin{tabular}{c l c c c c}
        \toprule
        \textbf{ID} &
        \textbf{Name} &
        \textbf{Material} &
        \textbf{Diameter [mm]} &
        \textbf{Density [kg/m]} &
        \textbf{End Weight [g]} \\
        \midrule
        1 & \#10 Sash Spot Cord   & Cotton & 9  & 0.040 & 18 \\
        2 & \#14 Spot Cord        & Cotton & 12 & 0.081 & 80 \\
        3 & Soft Braided          & Cotton & 15 & 0.076 & 80 \\
        4 & Shoe Lace             & Cotton & 7  & 0.014 & 5  \\
        5 & Thick Twisted         & Cotton & 25 & 0.139 & 50 \\
        6 & 3/8" Chain            & Steel  & 20 & 0.514 & 50 \\
        7 & 3/8" Surgical Tubing  & Latex  & 9  & 0.026 & 18 \\
        \bottomrule
    \end{tabular}
    \label{table:params}
\end{table*}
\subsection{Rope Parameters}
\label{sec:rope_params}
Rope types were selected to evaluate the algorithm robustness to diameter, material, and density. The end weight for each rope was chosen as the lightest weight that allowed the human demonstrator to verify the flying knot was achievable on that rope; the only demonstration used as the learning starting point is the rope 1 demonstration. A full list of parameters for each rope type used in evaluation can be found in \cref{table:params}. 

\subsection{Rope Dynamics Model}
\label{sec:dynamics_model}
We model the rope as a serial chain of $N$ point masses in maximal coordinates and simulate it with the first-order variational integrator described by \citet{lavalle_linear-time_2021}. In maximal coordinate dynamics formulations, each link of the rigid body mechanism is tracked in a global frame. Interactions between links such as joints or contacts are represented explicitly in an optimization problem solved at each timestep. Similarly, in our rope model, we do not track the joint angles between links but instead only track the global position and velocity of each link, then impose constraints to enforce the serial structure of the rope. Let $\mathbf{p}_k^{(i)}\in\mathbb{R}^3$ and $\mathbf{v}_k^{(i)}\in\mathbb{R}^3$ denote the position and velocity of link $i\in\{1,\dots,N\}$ at discrete timestep $k$. We stack all link positions and velocities as
\begin{align*}
    \mathbf{p}_k &:= \begin{bmatrix} {\mathbf{p}_k^{(1)}}^\top & \cdots & {\mathbf{p}_k^{(N)}}^\top \end{bmatrix}^\top \in \mathbb{R}^{3N}, \\
    \mathbf{v}_k &:= \begin{bmatrix} {\mathbf{v}_k^{(1)}}^\top & \cdots & {\mathbf{v}_k^{(N)}}^\top \end{bmatrix}^\top \in \mathbb{R}^{3N},
\end{align*}
and define the rope state (used throughout the paper) as $\state_k := \begin{bmatrix}\mathbf{p}_k^\top & \mathbf{v}_k^\top\end{bmatrix}^\top$. The integrator also introduces constraint multipliers $\boldsymbol{\lambda}_k$, and we denote the full maximal-coordinate variable by $\mathbf{z}_k := \begin{bmatrix}\mathbf{p}_k^\top & \mathbf{v}_k^\top & \boldsymbol{\lambda}_k^\top\end{bmatrix}^\top$. In \cref{eq:dynamics}, $\mathbf{z}_0$ denotes the initial rope configuration, with $\boldsymbol{\lambda}_0=\mathbf{0}$.

The rope is driven by the robot fingertip position $\mathbf{p}_{\mathrm{tip},k}\in\mathbb{R}^3$, which we treat as an input. The fingertip orientation is used to compute the bending stiffness and damping forces applied to the first link of the rope, capturing how the hand's orientation steers the rope at the grip. We enforce inextensibility via fixed-distance constraints between adjacent links of the rope and between the fingertip and the first link:
\begin{align*}
    g(\mathbf{p}_k, \mathbf{p}_{\mathrm{tip},k})
    &:=
    \begin{bmatrix}
        \|\mathbf{p}_k^{(1)} - \mathbf{p}_{\mathrm{tip},k}\|^2 - l^2 \\
        \|\mathbf{p}_k^{(2)} - \mathbf{p}_k^{(1)}\|^2 - l^2 \\
        \vdots \\
        \|\mathbf{p}_k^{(N)} - \mathbf{p}_k^{(N-1)}\|^2 - l^2
    \end{bmatrix}
    = \mathbf{0}.
\end{align*}
Let $\mathbf{G}(\mathbf{p}_k, \mathbf{p}_{\mathrm{tip},k}) := \frac{\partial g}{\partial \mathbf{p}}(\mathbf{p}_k, \mathbf{p}_{\mathrm{tip},k}) \in \mathbb{R}^{N\times 3N}$ be the constraint Jacobian, and let $\boldsymbol{\lambda}_k\in\mathbb{R}^{N}$ be the corresponding Lagrange multipliers, so that constraint forces on the links are $\mathbf{G}^\top \boldsymbol{\lambda}_k$.

The mass matrix is block diagonal,
\begin{align*}
    \mathbf{M}_{\mathrm{mass}} := \operatorname{diag}\!\left(m\mathbf{I}_{3(N-1)},\; m_e \mathbf{I}_3\right)\in\mathbb{R}^{3N\times 3N},
\end{align*}
and we collect gravity and internal bending forces (stiffness $k$ and damping $b$) into a single force vector $\mathbf{f}(\mathbf{p},\mathbf{v})\in\mathbb{R}^{3N}$. With these definitions, a single timestep of the variational integrator can be written as an implicit system in the unknowns $(\mathbf{p}_{k+1},\mathbf{v}_{k+1},\boldsymbol{\lambda}_{k+1})$:
\begin{align}
    \mathbf{0} &=
    \mathbf{p}_{k+1} - \mathbf{p}_k - \Delta t\, \mathbf{v}_{k+1},
    \label{eq:rope_vi_pos}\\
    \mathbf{0} &=
    \mathbf{M}_{\mathrm{mass}}(\mathbf{v}_{k+1}-\mathbf{v}_k)
    - \Delta t\,\mathbf{f}(\mathbf{p}_{k+1}, \mathbf{v}_{k+1})
    \nonumber\\
    &\quad
    - \Delta t\,\mathbf{G}(\mathbf{p}_{k+1}, \mathbf{p}_{\mathrm{tip},k+1})^\top \boldsymbol{\lambda}_{k+1},
    \label{eq:rope_vi_dyn}\\
    \mathbf{0} &= g(\mathbf{p}_{k+1}, \mathbf{p}_{\mathrm{tip},k+1}).
    \label{eq:rope_vi_con}
\end{align}
We solve \cref{eq:rope_vi_pos,eq:rope_vi_dyn,eq:rope_vi_con} with Newton's method at each timestep to roll out the rope trajectory, which defines the simulator $f$ used in \cref{eq:dynamics}.

To compute the local linear model used in the inverse-model QP, we differentiate an implicit system. Let $\tilde{\mathbf{z}}$ stack all per-timestep unknowns $(\mathbf{p}_k,\mathbf{v}_k,\boldsymbol{\lambda}_k)$ over the horizon and let $\tilde{\mathbf{u}}$ stack the driven fingertip positions $\mathbf{p}_{\mathrm{tip},k}$. Concatenating \cref{eq:rope_vi_pos,eq:rope_vi_dyn,eq:rope_vi_con} over time yields an implicit residual $\tilde f(\tilde{\mathbf{z}},\tilde{\mathbf{u}})=\mathbf{0}$. For a solution $(\tilde{\mathbf{z}}^\ast,\tilde{\mathbf{u}}^\ast)$, the implicit function theorem gives
\begin{align}
    \frac{\partial \tilde{\mathbf{z}}^\ast}{\partial \tilde{\mathbf{u}}}
    =
    -\left(\frac{\partial \tilde f}{\partial \tilde{\mathbf{z}}}\right)^{-1}
    \frac{\partial \tilde f}{\partial \tilde{\mathbf{u}}}.
    \label{eq:rope_ift}
\end{align}
In practice we do not form the inverse explicitly and instead solve linear systems with the KKT Jacobian $\frac{\partial \tilde f}{\partial \tilde{\mathbf{z}}}$. We implement the residual and its derivatives in Drake using automatic differentiation \cite{drake}, and apply the chain rule through the Bezier spline and forward kinematics to obtain the system-model linearization $\mathbf{M}$ in \cref{eq:dyn}.

\subsection{Command Parametrization}
\label{sec:bezier}
The Bezier trajectory $\mathcal{B}$ is parametrized by the 8 knot points $\knots \in \mathbb{R}^{10\times 8}$ equally spaced in the time interval from 0 to $T$. $\mathcal{B}$ is defined as follows, with $b_i$ representing the Bernstein basis functions:
\begin{align*}
\mathcal{B}(\knots, T)
&= \sum_{i=0}^{N} \knots_i \, b_i^{N}\!\left(\frac{t}{T}\right), \quad t \in [0, T] \\
b_i^{N}(s)
&= \binom{N}{i} s^{i} (1 - s)^{N-i}, \quad s \in [0,1].
\end{align*}

\subsection{Demonstration Timing Selection}
\label{sec:demoselect}
Given a full demonstration of the flying knot, we select a time window from $t_0$ to $t_f$ to limit demonstration motion to between the start of hand motion and the end of the follow-through motion. We select these times given a manually annotated rope collision time $t_c$ with the following search procedure:
\begin{enumerate}
    \item Select a temporary range $\tilde t_0$ and $\tilde t_f$ that excludes noisy data from the human picking up and placing the rope on the floor.
    \item Search for maximum hand velocity in the range $[\tilde t_0, \tilde t_f]$ and label as $t_{\text{peak}}$
    \item Step backwards in time from $t_{\text{peak}}$ and update $\tilde t_0$ to the first point when the hand velocity is near-zero (3\% of velocity at $t_{\text{peak}}$).
    \item Set $t_f=t_c + 35\text{ms}$  as a fixed follow-through time length.
    \item Integrate along the hand path motion between $\tilde t_0$ and $t_f$, then set $t_0$ to the time when 5\% of the total path length is traveled.
\end{enumerate}
$h(t)$ is then defined as 3D pose trajectory of the hand between $t_0$ and $t_f$. $\state^{\text{demo}}(t)$ is the rope trajectory from $t_0$ to $t_c$. Each demonstration type has a different execution speed and overall time length. 

\begin{table}[t]
\centering
\caption{Demonstration Tracking Parameters}
\label{tab:ikparams}
\begin{tabular}{l c}
\toprule
\textbf{Parameter} & \textbf{Value} \\
\midrule
$\mathbf{q_{\min}}$ & $[\;-6.28, -1.8, -6.28, -0.19, -6.28, -1.69, -6.28\;]$ \\
$\mathbf{q_{\max}}$ & $[\;6.28, 1.9, 6.28, 3.92, 6.28, 3.14, 6.28\;]$ \\
$\mathbf{\dot q_{\min}}$ & $-[3.14, 3.14, 3.14, 3.14, 3.14, 3.14, 3.14]$ \\
$\mathbf{\dot q_{\max}}$ & $[3.14, 3.14, 3.14, 3.14, 3.14, 3.14, 3.14]$ \\
$\mathbf{\ddot q_{\min}}$ & $-[100, 100, 100, 100, 100, 100, 100]$ \\
$\mathbf{\ddot q_{\max}}$ & $[100, 100, 100, 100, 100, 100, 100]$ \\
$w_j$ & $5.0 \times 10^{-7}$ \\
$w_p$ & $\;10\;$ \\
$w_R$ & $\;0.2\;$ \\
$w_v$ & $\;0.5\;$ \\
$z_{\min}$ & $\;1.2\;$ \\
\bottomrule
\end{tabular}
\end{table}
\subsection{Demonstration Tracking}
\label{sec:track_hand}
The first 7 rows of $\mathbf{q}(t)$ correspond to the robot joint trajectory and the remaining 3 rows correspond to the robot base location. $\mathbf{e}_h(t)$ is the end-effector tracking error with respect to the desired fingertip trajectory $h(t)$ weighted by the cost matrix $\mathbf{W}_h$:
\begin{align*}
\mathbf{e}_h(t) :=
\begin{bmatrix}
\mathbf{p}_{\mathrm{tip}}(t)-\mathbf{p}_h(t)\\
\operatorname{Log}\!\bigl(\mathbf{R}_h(t)^\top \mathbf{R}_{\mathrm{tip}}(t)\bigr)\\
\dot{\mathbf{p}}_{\mathrm{tip}}(t)-\dot{\mathbf{p}}_h(t)
\end{bmatrix}\\
\mathbf{W}_h := \operatorname{diag}(w_p\mathbf{I}_3,\;w_R\mathbf{I}_3,\;w_v\mathbf{I}_3),
\end{align*}
with $\mathbf{p}_{\mathrm{tip}}(t)$, $\mathbf{R}_{\mathrm{tip}}(t)$, and $\dot{\mathbf{p}}_{\mathrm{tip}}(t)$ given by the robot forward kinematics $\mathcal{K}$. Values for $w_p, w_R, w_v$ can be found in \cref{tab:ikparams}. The joint dynamics limits in \cref{tab:ikparams} and \cref{tab:invmodelparams} are the same. The demonstration-tracking objective tracks the hand motion without any critical-point weighting, so the cost terms are applied equally throughout the hand motion.

The $z_{\mathrm{tip}}$ constraint specifies that the initial configuration of the robot at $t_0$ must have enough distance for the rope to dangle without hitting the floor.

\fi 

\end{document}